%% file: main.tex
\documentclass[10pt,twocolumn,letterpaper]{article}

\usepackage{cvpr}              

\input{preamble}

\definecolor{cvprblue}{rgb}{0.21,0.49,0.74}
\usepackage[pagebackref,breaklinks,colorlinks,citecolor=cvprblue]{hyperref}
\usepackage{multirow}
\def\confName{CVPR}
\def\confYear{2024}

\title{ObjectNLQ @ Ego4D Episodic Memory Challenge 2024}

\author{
Yisen Feng$^{1}$, Haoyu Zhang$^{1\,2}$, Yuquan Xie$^{1}$, Zaijing Li$^{1\,2}$, Meng Liu$^{3}$, Liqiang Nie$^{1}$\\
$^1$Harbin Institute of Technology (Shenzhen) \qquad  $^2$Peng Cheng Laboratory    \\$^3$Shandong Jianzhu University\\
{\tt\small \{23S051028,23S051007\}@stu.hit.edu.cn;}\\
{\tt\small \{zhang.hy.2019, lzj14011, mengliu.sdu, nieliqiang\}@gmail.com} 
}

\begin{document}
\maketitle
\input{sec/0_abstract}
\input{sec/1_intro}

\input{sec/2_methodology}
\input{sec/3_experiment}

\input{sec/4_conclusion}
{
    \small
    \bibliographystyle{ieeenat_fullname}
    \bibliography{main}
}


\end{document}

%% file: preamble.tex
\usepackage[dvipsnames]{xcolor}


%% file: sec/0_abstract.tex
\begin{abstract}
    In this report, we present our approach for the Natural Language Query track and Goal Step track of the Ego4D Episodic Memory Benchmark at CVPR 2024. Both challenges require the localization of actions within long video sequences using textual queries. To enhance localization accuracy, our method not only processes the temporal information of videos but also identifies fine-grained objects spatially within the frames. To this end, we introduce a novel approach, termed ObjectNLQ, which incorporates an object branch to augment the video representation with detailed object information, thereby improving grounding efficiency. 
    ObjectNLQ achieves a mean R@1 of 23.15, ranking 2nd in the Natural Language Queries Challenge, and gains 33.00 in terms of the metric R@1, IoU=0.3, ranking 3rd in the Goal Step Challenge. Our code will be released at \url{https://github.com/Yisen-Feng/ObjectNLQ}.
\end{abstract}

%% file: sec/1_intro.tex
\section{Introduction}

\begin{figure}[t]
  \centering
  \includegraphics[width=0.9\linewidth]{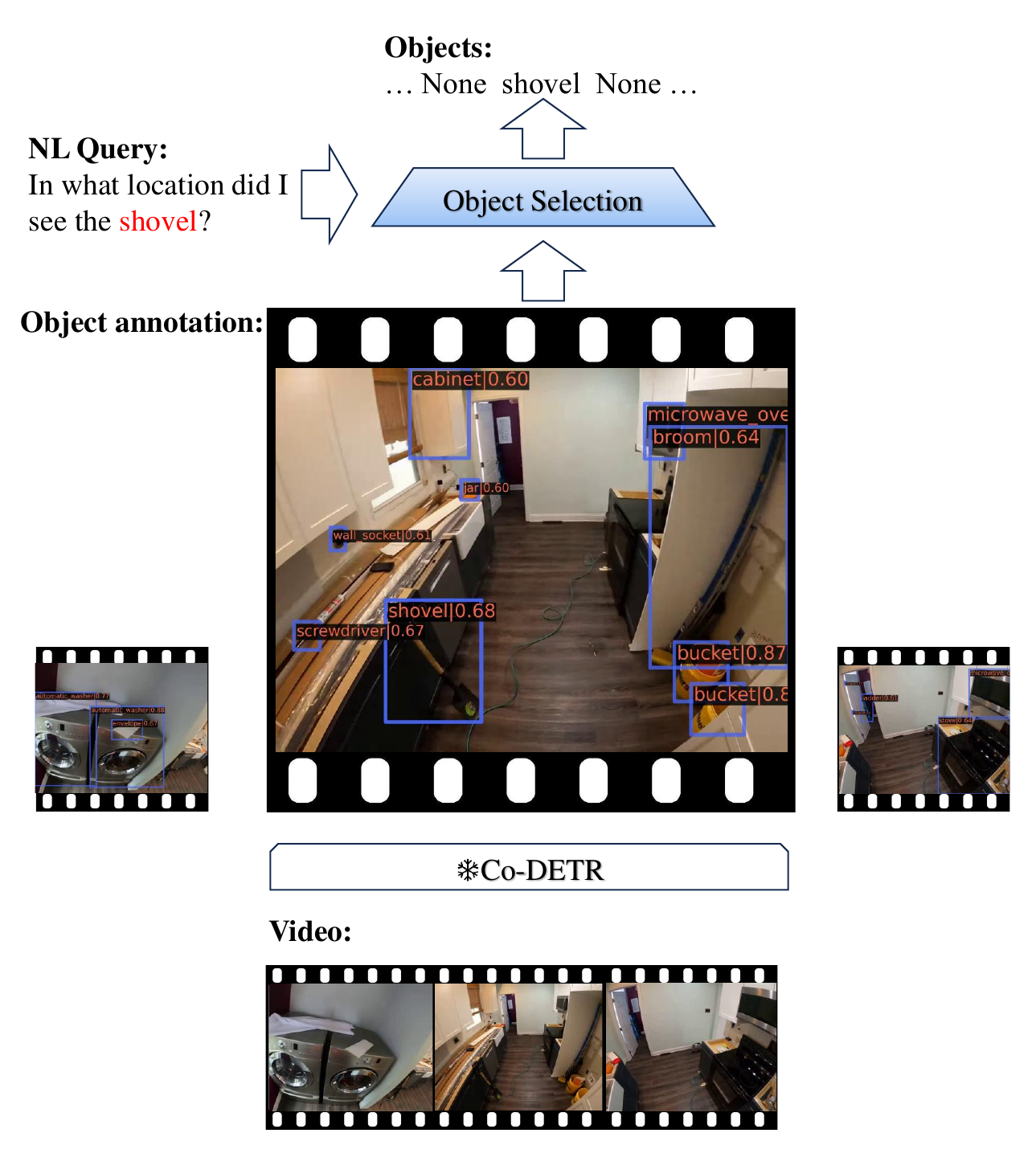}
  \vspace{-2ex}
  \caption{An illustration of object extraction.}
  \vspace{-2ex}
  \label{figure1}
\end{figure}

In the Ego4D~\cite{grauman2022} Natural Language Query (NLQ) Challenge, participants are provided with an egocentric video and a natural language question. The objective is to accurately localize the video segment that contains the answer to the question~\cite{pmlr-v235-zhang24aj,10.1145/3474085.3475234}. Conversely, the Goal Step Challenge \cite{song2024ego4d} requires localizing a video segment that corresponds to a natural language description of the step. Both challenges necessitate precise identification of relevant video content based on textual queries.

Existing methods have made significant advancements in several key areas: 1) Developing a robust feature extraction backbone through pretraining~\cite{Lin2022,Chen2022}. These methods finetune a pre-trained multimodal video model on the Ego4D dataset to enhance feature representation for egocentric videos. 
2) Implementing efficient data augmentation strategies to expand the scale of the dataset~\cite{Ramakrishnan2023}. This enhances the performance of downstream models by effectively increasing data variability. 3) Designing tailored grounding models for the NLQ task~\cite{Lin2022,Liu2022a,Hou2023a,Mo2022,hou2023}. Some approaches optimize the framework by reducing the number of frames processed, thereby streamlining video moment localization. Others~\cite{Mo2022,hou2023} leverage the sophisticated architectures of point prediction and feature pyramids from ActionFormer~\cite{Zhang2022} to process the entire video directly, aiming for more precise localization.
\begin{figure*}[t]
  \centering
  \includegraphics[width=0.75\linewidth]{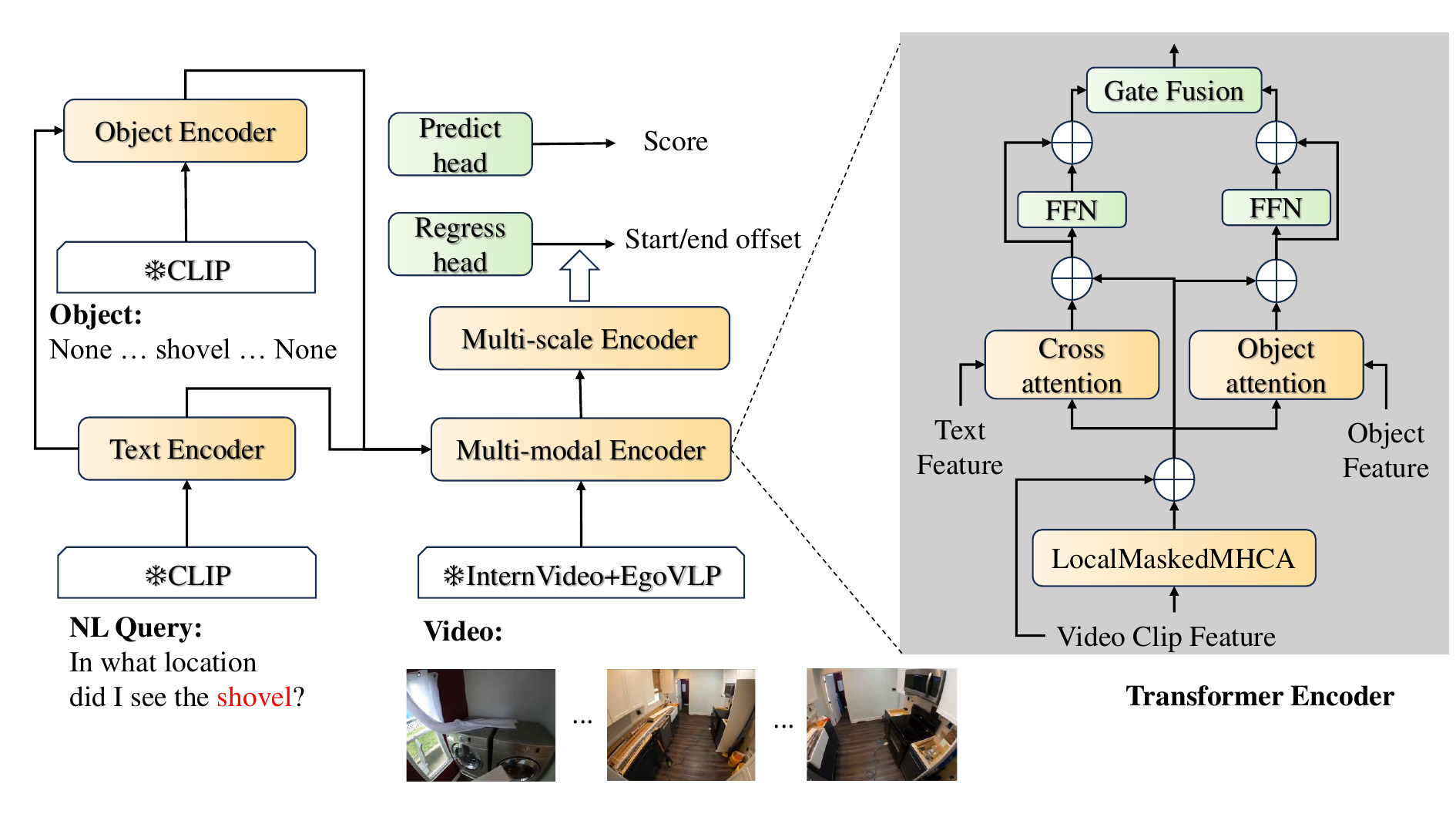}
  \vspace{-2ex}
  \caption{The framework of our proposed localization model.}
  \label{figure2}
\end{figure*}

Among the above three methods, the extraction of video features is crucial, and the quality of feature extraction will directly determine whether the model can recognize the specific information in the video. However, existing pretraining methods commonly finetune the backbone using contrastive learning between captions and video clips, which means the video features depend entirely on the granularity and focus of caption information. Moreover, most of the captions in the training data focus on the behavior of human, while the NLQ task concentrates on building an AI helper for finding things that people often neglect. This mismatch leads to the lack of fine-grained object information in the input.

Based on this observation, we propose a novel method for object extraction and design a novel grounding model that can fully use the object annotation. Through an object detection model, we obtain the fine-grained object information in the image. Then we use a parallel attention mechanism to extract object information and add it to video frame features, to reasonably complete the NLQ task.

%% file: sec/2_methodology.tex
\section{Methodology}
This section introduces the comprehensive methodology employed in our approach. The extraction of video, text, and object features is detailed in Section \ref{sec:feature extraction}. The architecture and functioning of our grounding model are described in Section \ref{sec:Structure}. We discuss the loss function used for training the model in Section \ref{sec:Loss}. 


\begin{table*}[ht]
  \caption{Performance comparison on NLQ Challenge.}
  \vspace{-2ex}
  \label{table1}
  \centering
  \begin{tabular}{ccccccc}
    \toprule
    \multirow{2}{*}{Entry}&\multirow{2}{*}{Rank}  & \multicolumn{3}{c}{R@1(\%)}  & \multicolumn{2}{c}{R@5(\%)}        \\
     &   &0.3 & 0.5 & Mean & 0.3 & 0.5 \\
    \midrule
     GroundNLQ\cite{hou2023}&-  & 25.67 & 18.18 & 21.93 & 42.06 & 29.8  \\
     
     EgoVideo&1 & \textbf{28.05} & \textbf{19.31} & \textbf{23.68} &	\textbf{44.16} & \textbf{31.37}  \\
     \midrule
     ObjectNLQ&2 & 27.02 & 19.28 & 23.15 &43.66 & 30.87 \\
    \bottomrule
  \end{tabular}
\end{table*}

\subsection{Feature Extraction}
\label{sec:feature extraction}
Following  \cite{hou2023}, we employ a concatenated feature comprising InternVideo \cite{Chen2022} and EgoVLP \cite{Lin2022} for video representations, while utilizing CLIP (ViT-L/14) \cite{Radford2021LearningTV} for textual feature extraction. For object representation, as illustrated in Figure \ref{figure1}, we utilize the Co-DETR \cite{zong2023detrs} object detector, which is pretrained on the LVIS v1.0 dataset \cite{Gupta2019}, to extract object annotations from the video. Subsequently, we employ CLIP (ViT-L/14) \cite{Radford2021LearningTV} to extract the textual features of the object classes that correspond to the given query.


\subsection{Structure}
\label{sec:Structure}

As depicted in Figure \ref{figure2}, our grounding model incorporates object information\footnote{For the Goal Step task, since the text description closely resembles the captions used in the pretraining data, we opt not to incorporate additional object data. Instead, we apply the pure GroundNLQ \cite{hou2023} architecture to this task.} and consists of several key components: a text encoder, an object encoder, a multi-modal encoder, a multi-scale encoder, and a prediction head. In alignment with \cite{hou2023}, we employ the same text encoder, multi-scale transformer encoder, and prediction heads.

To improve the comprehension of fine-grained object information, we have meticulously designed an object encoder. This encoder is tailored to filter and highlight object information relevant to the query. Additionally, we integrate an object branch into the multi-modal encoder, thereby significantly enhancing the representation of video features. 
\paragraph{Object Encoder.} 
Our object encoder comprises four transformer encoder blocks. Each block features a text cross-attention layer \cite{10239469} followed by a feed-forward network (FFN). In this architecture, the object feature acts as the query in the text cross-attention layer, with the text feature serving as both the key and value. This setup enables our model to thoroughly extract and utilize object data pertinent to the specific query, effectively minimizing confusion about objects from different categories within the same frame. 

\paragraph{Multi-Modal Encoder.}
Our multi-modal encoder comprises four transformer encoder blocks, each featuring a sliding window multi-head self-attention (MHA) layer, a text branch, an object branch, and a gate fusion module. Within each branch, a cross-attention layer and an FFN are employed. Subsequently, the outputs from the two branches are combined using a gate fusion technique.

Specifically, the left branch aligns with the design of the multi-modal encoder proposed in \cite{hou2023}, where the text feature acts as both key and value. The right branch, however, utilizes the video feature as the query while employing the object feature as the key and value. The integration is achieved through a gate fusion process, where the weights for merging are computed by a multi-layer perceptron (MLP). This innovative design seamlessly combines object data with video features, maintaining the integrity of the original interactions between text and video content, thus enhancing the overall model's capability to interpret and utilize multimodal data.


\subsection{Loss Function}
\label{sec:Loss}
Following \cite{hou2023}, our model employs binary classification loss for the classification tasks and IoU (Intersection over Union) regression loss for regression tasks. Additionally, to enhance the model's sensitivity to action detection and localization, we incorporate a learnable Gaussian weighting on these losses, similar to the approach described in \cite{Shao2023}.



%% file: sec/3_experiment.tex
\section{Experiment}
\subsection{Implementation Details}

Video clip features are extracted with a stride of 16 frames, and the dimension is set to 384. We utilize four heads for multi-head attention. During training, configurations differ between tasks: for the NLQ task, we use a mini-batch size of 4 and a learning rate of 1e-4; for the Goal Step task, the mini-batch size is increased to 8 with a learning rate of 2e-4. We employ a cosine decay strategy for learning rate adjustment, with a warm-up period of four epochs and a total training duration of ten epochs. The training for the NLQ task is conducted on a single L20 GPU over approximately four hours, while the Goal Step task requires four L20 GPUs for around ten hours of training. For model initialization, we use pretrained parameters to ensure a robust starting point. The original parameters of GroundNLQ are initialized with weights from a model pretrained on narration data, which has been augmented using the strategy of \cite{Ramakrishnan2023}. Within the object encoder, the text cross-attention layers and the FFN are initialized using the MHA layer and the FFN from the text encoder. Similarly, the object branch in the multi-modal encoder is initialized using the parameters from the text branch, ensuring consistency and leveraging pretrained efficiencies across the model architecture.

To maximize the use of available data, both the training and validation splits are divided into five folds. This data is used to train multiple models for an ensemble approach, enhancing robustness and accuracy. During inference, SoftNMS is employed for deduplication to improve the precision of localization~\cite{10.1145/3503161.3548020}.

\subsection{Performance Comparison}

\begin{table}[t]
  \caption{Performance comparison on the Goal Step Challenge.}
  \vspace{-2ex}
  \label{table2}
  \centering
  \begin{tabular}{cccc}
    \toprule
    \multirow{2}{*}{Entry}&\multirow{2}{*}{Rank}  & \multicolumn{2}{c}{R@1(\%)}   \\
      &   &0.3 & 0.5  \\
    \midrule
    VSLNet(Omnivore) \cite{Zhang2020}&6 &19.04&12.04\\
    LorCar&5&24.75&16.55\\
    flyfishing&4&29.69&18.99\\
     EgoVideo&2  & 34.06 & \textbf{26.97}  \\
     VSLNet Prob&1 & \textbf{35.18} & 20.49 \\
     \midrule
     ObjectNLQ&3 & 33.00 & 26.37  \\
    \bottomrule
  \end{tabular}
\end{table}
Table~\ref{table1} presents the leaderboard results of the NLQ Challenge. Our ensemble method achieves a mean R@1 score of 23.15\%, which is 1.22\% higher than the 2023 champion and secured us a 2nd-place ranking. Table \ref{table2} details the leaderboard for the Goal Step Challenge, where our ensemble method attains an R@1 score of 33.00\% at an IoU of 0.3. This score is 13.96\% points higher than the baseline and placed us third. These results underscore the effectiveness of our method in both challenges.

\subsection{Ablation Study}
\begin{table}[t]
  \caption{Ablation study on the validation set of NLQ.}
  \vspace{-2ex}
  \label{table3}
  \centering
  \begin{tabular}{cccccc}
    \toprule
    \multirow{2}{*}{Entry}  & \multicolumn{3}{c}{R@1(\%)}  & \multicolumn{2}{c}{R@5(\%)}        \\
     &   0.3 & 0.5 & Mean & 0.3 & 0.5 \\
    \midrule
    SA+CA& 28.05&19.53&23.79&56.52&41.63\\
    SOA&28.19&19.42&23.81&56.11&41.15\\ 
     w/o ASL \cite{Shao2023} & 28.43 & 19.82 & 24.13 & 55.95 & 41.56  \\
     ObjectNLQ & 28.43 & 19.95 & 24.19 &56.06 & 42.09 \\
    \bottomrule
  \end{tabular}
\end{table}
In our ablation experiments conducted on the validation split, the results are systematically detailed in Table \ref{table3} to assess various components of our model:
1) At the top of Table~\ref{table3}, we examine modifications to the object encoder. In this variant, we build the object encoder with an additional self MHA layer (SA), a text cross-attention layer (CA), and an FFN. Interestingly, the performance slightly decreases, suggesting that excessive associations among objects within the same frame might be detrimental.
2) At the second line of Table~\ref{table3}, we explore the design of the video encoder. The ablation setup, Sequential Object Attention (SOA), consists of an MHA layer, followed by a text cross-attention layer, an object cross-attention layer, and an FFN. The observed decline in performance with this configuration reinforces the efficacy of our original video encoder design, indicating that simpler sequences may be more effective.
3) At the third line of Table~\ref{table3}, we ablate the use of a Gaussian weight on the loss function. The slight decrease in performance upon removing this feature confirms the utility of Action Sensitivity Learning (ASL) \cite{Shao2023}, which enhances the model's focus on relevant aspects of the data for improved performance.

\subsection{Case Analysis}
\begin{figure}
    \centering
    \subfloat[Success example]{
        \label{fig: nlq success example}		
      \includegraphics[width=0.9\linewidth]{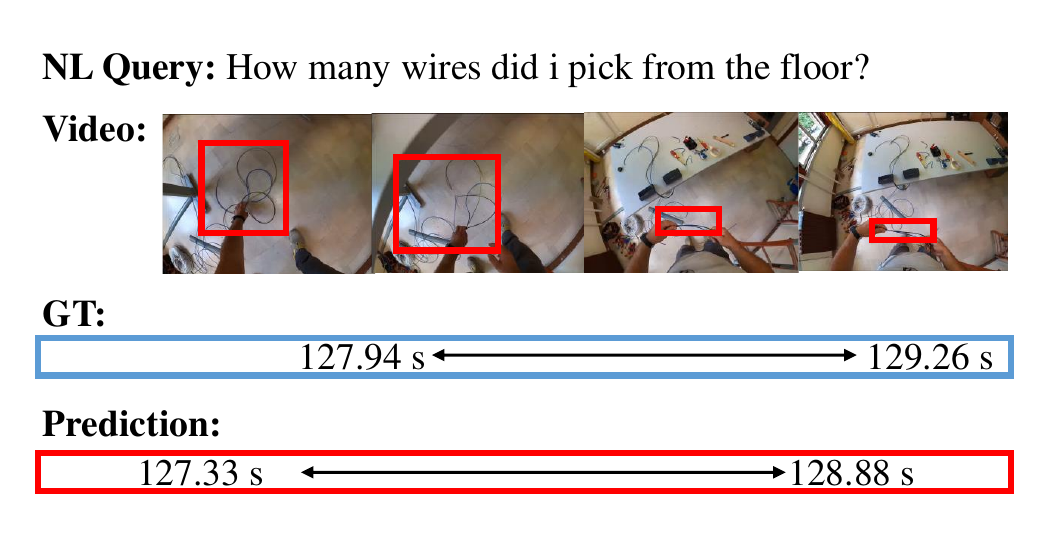}}\\
    \subfloat[Failure example]{
        \label{fig: nlq failure example}
    		\includegraphics[width=0.9\linewidth]{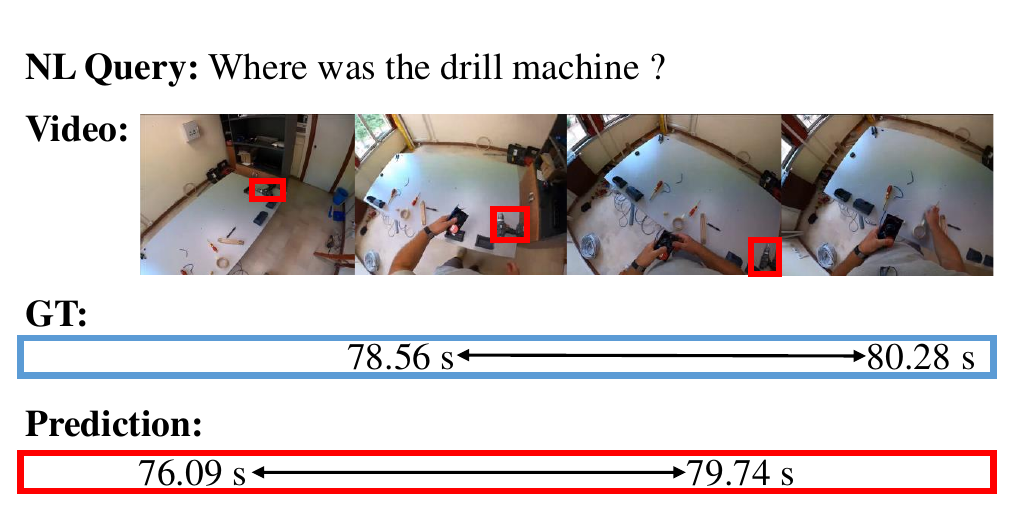}}
      \vspace{-2ex}
    \caption{Two examples of ObjectNLQ on the validation set of NLQ.}
    \vspace{-2ex}
    \label{figure3}
\end{figure}
\begin{figure}
    \centering
    \subfloat[Success example]{	
    \label{fig: goalstep success example}
      \includegraphics[width=0.9\linewidth]{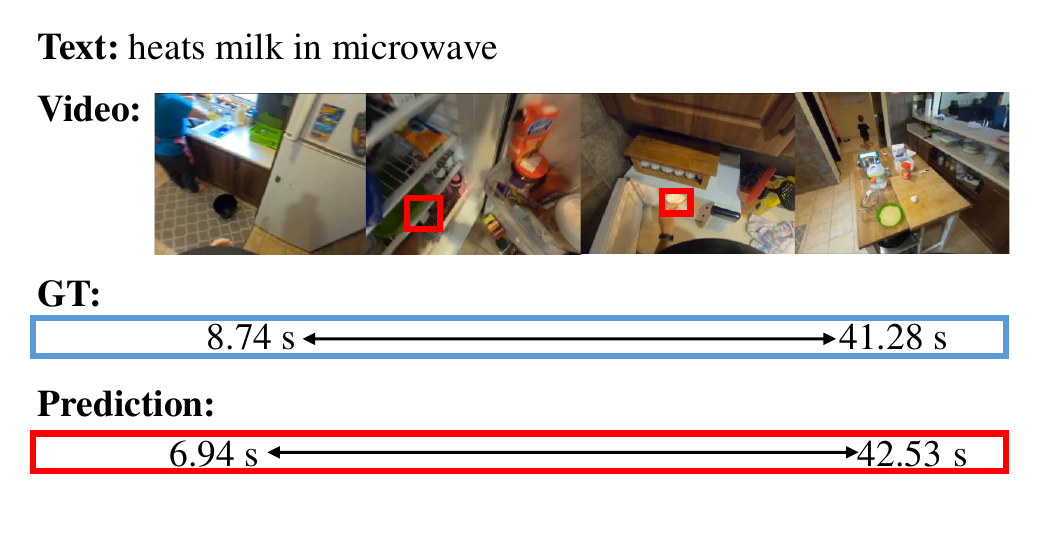}}\\
    \subfloat[Failure example]{
    \label{fig: goalstep failure example}
    		\includegraphics[width=0.9\linewidth]{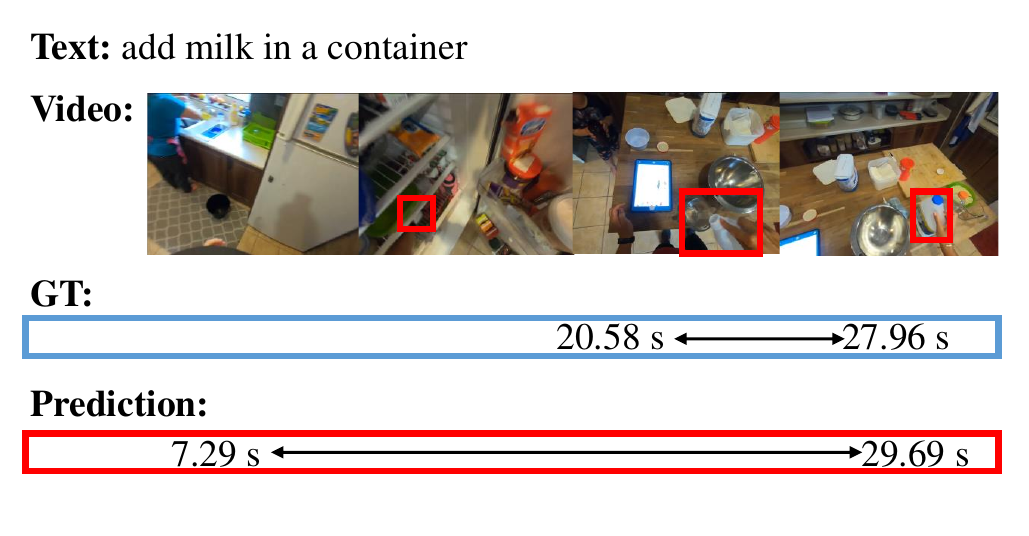}}
      \vspace{-2ex}
    \caption{Two examples of ObjectNLQ on the validation set of Goal Step.}
    \vspace{-2ex}
    \label{figure4}
\end{figure}
In Figure \ref{fig: nlq success example}, ObjectNLQ accurately locates~\cite{zhang2023uncovering} the ground truth of the query, effectively integrating both the specified object and the action ``pick'' mentioned in the query. This successful prediction exemplifies the model’s ability to simultaneously process added object information and inherent video action details, clearly highlighting the advantages of our architectural design.
Conversely, Figure \ref{fig: nlq failure example} presents a case where our model did not perform as expected. While it correctly identifies the object in the scene, it fails to accurately localize the answer, particularly because the ground truth required the drill machine to be centered in the frame. This failure suggests that the model may not adequately understand or prioritize human attention cues, indicating a potential area for further refinement to enhance its performance in complex scenarios.

Figure \ref{fig: goalstep success example} shows a successful example in Goal Step. Our model exactly locates the target of the text description, which contains a sequence of actions including taking milk from a fridge, pouring milk into a container, and heating milk with a microwave. This case reflects our model's ability to identify the location intent as well as granularity.
However, an example of underperformance is demonstrated in Figure \ref{fig: goalstep failure example}. Our model incorrectly adds the removal of milk from the refrigerator to the localization results, resulting in inaccurate localization. This reflects that our model may not be able to fully understand the semantics of fine-grained actions, and improving the model's ability to understand and segment long action sequences may further improve its performance.

%% file: sec/4_conclusion.tex
\section{Conclusion}


In this report, we present our approach for the NLQ track and Goal Step track of the Ego4D Challenge at CVPR 2024. Our primary contribution lies in leveraging existing object detection models to extract detailed object information from video frames and integrating this data into an effective model framework to enrich video representation with this additional object detail. Employing fine-grained object features, our method improves upon the foundational GroundNLQ model in the NLQ Challenge, demonstrating enhanced performance.